\documentclass[runningheads]{llncs}
\usepackage[noend]{algpseudocode}
\usepackage{amsmath}
\usepackage[linesnumbered,ruled]{algorithm2e}
\usepackage{graphicx}
\usepackage{subcaption}
\graphicspath{{Figures/}}
\begin{document}
\title{Algorithmic Factors Influencing \\Bias in Machine Learning}
%
%

\author{William Blanzeisky \and P\'{a}draig Cunningham}
	\institute{School of Computer Science\\
	University College Dublin\\
	Dublin 4, Ireland\\
}

\titlerunning{Algorithmic Bias}

\authorrunning{Blanzeisky \& Cunningham}

\maketitle              
\begin{abstract}

It is fair to say that many of the prominent examples of bias in Machine Learning (ML) arise from bias that is there in the training data. In fact, some would argue that supervised ML algorithms \emph{cannot} be biased, they reflect the data on which they are trained. In this paper we demonstrate how ML algorithms can misrepresent the training data through \emph{underestimation}. We show how irreducible error, regularization and feature and class imbalance can contribute to this underestimation. The paper concludes with a demonstration of how the careful management of synthetic counterfactuals can ameliorate the impact of this underestimation bias.


\end{abstract}

\section{Introduction}

As the applications of Machine Learning (ML) systems become ubiquitous in many aspects of humans life, there is an undeniable appeal for discrimination-free machine learning. An algorithm is considered to be discriminatory if it systematically disadvantages people belonging to certain categories or groups, instead of relying solely on individual merits \cite{Zliobaite2017b}. Despite the increasing number of related works over the past few years, the amount of anecdotal evidence of algorithmic bias is still growing \cite{10.1145/3442188.3445922}. 

There are two main sources of algorithmic bias, it can be due to the data or the algorithm. When it is due to the data, it is sometimes euphemistically called \emph{negative legacy}, when it is due to the algorithm it is called \emph{underestimation}. \cite{cunningham2020algorithmic,kamishima2012fairness}.  Negative legacy may be due to labeling errors or poor sampling; however, it is likely to reflect discriminatory practices in the past. 
Underestimation occurs when the algorithm focuses on strong signals in the data thereby missing more subtle phenomena.  Algorithmic aspects of how ML algorithms can accentuate existing bias are not yet well-understood. Hence, the focus of this paper is on underestimation.

Our central hypothesis is that underestimation occurs when an algorithm \emph{underfits} the training data due to a combination of  limitations in training data and model capacity issues. This leads us to a series of sub-hypotheses -- that the following factors can contribute to underestimation:

\begin{itemize}
    \item Irreducible error (Bayes error)
    \item Regularization mechanisms
    \item Class imbalance
    \item Under-represented categories
\end{itemize}
These hypotheses are presented in more detail in section \ref{sec:factors}. Before that the relevant background research is reviewed in section \ref{sec:background}. In section \ref{sec:syn_data} these hypotheses are tested on synthetic data. In section \ref{sec:fix} we show that the addition of synthetic counterfactuals to the training data can reduce the impact of underestimation but there is clearly a need to manage the number of counterfactuals. The paper concludes in section \ref{sec:real_data} with an assessment of how this repair strategy works on real datasets. 

\section{Background}\label{sec:background}








The issues around bias and fairness in ML research has received a lot of attention in recent years. Several notions of what constitutes ``fair" in ML have been proposed \cite{dunkelau2019fairness}. Despite this, there is still no unified consensus of what the best fairness notion shall be. In fact, recent works show that some of these notions suffer from significant statistical limitations, or even might perversely harm the very groups they were designed to protect \cite{corbettdavies2018measure}. In general, an ML model is considered fair if it is not inclined to award the desirable outcome $Y = 1$ (e.g. loan approval/job offers) only to one side of sensitive category $S \ne 1$ (e.g. gender/race). 

The fairness literature has mainly focused on implementing new methods to ensure fairness without explicitly considering the source of bias \cite{caton2020fairness}. Recent efforts in rectifying algorithmic bias include: transforming the dataset to remove discrimination before feeding it into a ML model (pre-processing) \cite{article}, modifying a specific algorithm's loss function to account for fairness (in-processing) \cite{kamishima2012fairness}, or transforming the ML model's output to ensure fairness (post-processing) \cite{10.5555/3157382.3157469} . 

Many argue that algorithmic bias is purely caused by negative legacy and that the algorithm models the data correctly (no underestimation). However, recent work shows that the algorithm itself could amplify existing bias \cite{cunningham2020algorithmic,hooker2020characterising}. Of particular interest in this regard is research on model pruning in deep neural networks \cite{hooker2020characterising,hooker2020compressed}. Since the 1990s there has been research on how a significant proportion of the weights in a neural network can be pruned with minimal impact on model accuracy. Recent work at Google Brain by Hooker \emph{et al.} \cite{hooker2020characterising,hooker2020compressed} shows that this pruning can have a significant impact on underestimation. This phenomenon is directly related to the regularization issue we demonstrate in section \ref{sec:imp_Reg}.

Although most examples of bias in ML systems occur due to negative legacy, we argue that it is important to understand  how specific algorithm mechanisms can introduce or at least accentuate bias. Understanding how algorithms can become discriminatory on the algorithmic level could help researchers in developing better and more general strategies to ensure fairness in ML algorithms. We believe that some aspects of ML training, such as choices around model architectures, hyper-parameters and optimization criteria, could influence algorithmic bias. 

\subsection{Quantifying Bias}

Disparate Impact ($\mathrm{DI}_S$) is one of the accepted definitions of unfairness \cite{feldman2015certifying}:
\begin{equation}\label{eqn:DI}
   \mathrm{DI}_S \leftarrow \frac{P[\hat{Y}= 1 | S \ne 1]}{P[\hat{Y} = 1 \vert S = 1]} < \tau 
\end{equation}
It is the ratio of desirable outcomes $\hat{Y}$ predicted for the sensitive minority $S \ne 1$ compared with that for the majority $S = 1$. $\tau = 0.8$ is the 80\% rule, i.e. proportion of desirable outcomes for the minority should be within 80\% of those for the majority. 

When the focus is on bias due to the algorithm only we can define an \emph{underestimation score} ($\mathrm{US}_{S}$) in line with $\mathrm{DI}_S$:
\begin{equation}\label{eqn:US_S1}
    \mathrm{US}_{S} \leftarrow \frac{P[\hat{Y}= 1 | S \ne 1]}{P[Y = 1 | S \ne 1]} 
\end{equation}
This is the ratio of desirable outcomes predicted by the classifier for the sensitive minority compared with what is actually present in the data. 
If $\mathrm{US}_{S} <1$ the classifier is under-predicting  desirable outcomes for the minority. It is also important to note that $\mathrm{US}_{S} =1$ does not necessarily mean that the classifier is not biased against the minority group (i.e. poor $\mathrm{DI}_S$) score. It simply means that the algorithm does not underestimate the predictions of desirable outcome for the minority group.

\section{Factors Contributing to Underestimation} \label{sec:factors} 
The definition of disparate impact emphasises fairness for all subgroups -- see equation \ref{eqn:DI}. This definition is independent of any data, indeed disparate impact is likely to be caused by biased historic data. By contrast the definition of underestimation depends on test data (equation \ref{eqn:US_S1}) -- an outcome is under-predicted for a category. Our hypothesis is that this will occur when a model underfits the data. We identify four factors that can contribute to this. 

\subsubsection{Irreducible Error:} In supervised ML the assumption is that the outcome variable $Y$ is determined by the inputs to the model. Normally the inputs do not completely determine the outcome (there are hidden factors involved) so even the best model will have some error. This is the irreducible error, sometimes referred to as Bayes error \cite{tumer1996estimating}. If the irreducible error is low the best models will be able to fit well to the training data and still generalise well to unseen data. If the irreducible error is high ($\sim25\%$) then models that can generalise well will need to be relatively simple, i.e. they will underfit the training data. We show in section \ref{sec:imp_IrrErr} that underestimation directly correlates with irreducible error. 

\subsubsection{Regularization Mechanisms:}  
It is standard practice in ML to use regularization methods to reduce generalization error. These methods are used to control the model capacity to gather relevant information from the training set. The objective of ensuring that the model does not overfit or underfit this information is crucial  in ML because it is tied to the model's ability to generalize well on unseen data.  There have been many proposed regularization techniques in the ML literature: lasso, ridge, elastic-net, dropout, early-stopping, etc \cite{doi:10.1142/S0129065791000352,doi:10.1080/00401706.1970.10488634}. The main idea behind these methods is to control the model's complexity: in some circumstances, this is done by adding a penalty term in the loss function, or by limiting the number of nodes in the hidden layer. In regression settings, these methods are used to mitigate over-fitting by penalizing the effect of each predictor in explaining the target variable (reduce variance). Thus, excessive regularization will likely cause a model to underestimate predictions for the minority class. Since regularization directly correlates with bias and variance of a model, we show that it also influences underestimation (see section \ref{sec:imp_Reg}).  

\subsubsection{Class Imbalance:} Class imbalance refers to a classification problem where the number of observations in the data set differs for each class. The potential for this to lead to bias has been known for some time. It is known that the model predictions can accentuate the bias: if the minority class represents 30\% of the training data the model is likely to predict $ < 30\%$ for that class \cite{kubat1998machine}
\cite{mac2002problem}. Training an ML model without addressing this issue will make it more challenging for a model to learn the characteristics of examples from the rare events, resulting in predictions that underestimate the overall minority class (i.e. biased towards predicting the majority) \cite{4667275}. Many remediation strategies have been proposed to combat these issues, such as: sampling, cost-sensitive learning, etc \cite{weiss2004imbalance}. In section 4, we will see the varying impact of class imbalance on underestimation. 

\subsubsection{Under-represented Categories:}
One of the most common causes of bias in ML is the lack of observations for under-represented groups/categories. For example, given a protected attribute/feature such as race or gender, if a model is trained on 5000 observations in which 80\% belongs to the majority and only 20\% represents the minority group, the model will likely struggle to learn as effectively from the minority examples. It is important to distinguish between this and the class imbalance issue highlighted above. This effect is aggravated when the irreducible error is high (the model needs to be simple).  Although the question of directly using protected attributes in ML training remains open \cite{Zliobaite2016}, we show in section \ref{sec:imp_Imb} that the level of under-representation directly correlates with underestimation.

\section{Underestimation on Synthetic Data} \label{sec:syn_data}

In this section, we present results of the experiments based on the hypothesis presented in section \ref{sec:factors}. These baseline results are demonstrated on a synthetic data model that is widely used in research on bias on ML \cite{adler2016auditing}. 

\begin{figure}[t]
    \centering
    \includegraphics[width=0.65\textwidth]{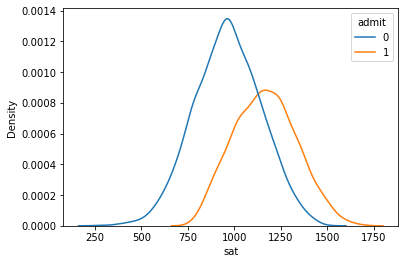}
    \caption{Probability density functions of the synthetic data where the blue and orange curve shows the distribution of SAT scores for admitted and rejected applicants respectively.}
    \label{fig:kdeplots}
\end{figure}

\subsection{Data Description}

This dataset consists of 5000 observations, each represents an applicant's statistics, which include IQ score, SAT score and a sensitive attribute (Favored, or Discriminated) \cite{adler2016auditing}. The target variable, admit, indicates whether or not a student is admitted. The synthetic data generated has the following key properties: (i) The observations are evenly distributed (50\% for both favored and discriminated group); (ii) The IQ scores are randomly sampled between 80 and 120; (iii) To reflect some correlation, the SAT scores are stochastic but dependent on the sensitive attribute and IQ; (iv) The admit outcome is also stochastic but correlated with the SAT score; and (v) since students in the favored group on average have higher SAT score than those that are in discriminated group, favored students have been admitted at a higher rate than discriminated students. 

70\% of these observations are used for training, and the remaining 30\% are reserved for model testing. We run  experiments on seven classifiers implemented in \textsf{scikit-learn}\footnote{\url{https://scikit-learn.org/}}. These classifiers include tree-based classifiers (Decision Tree, Gradient Boost and Random Forest), and four other classifiers; $k$-Nearest Neighbor, Naive Bayes, Logistic Regression \& Neural Networks. Taking into account the stochastic process of generating the synthetic sensitive attributes, each experiments were repeated twenty times and the median underestimation on the test set is obtained. 

\begin{figure}
\centering
\begin{subfigure}[b]{.45\linewidth}
\includegraphics[width=\linewidth]{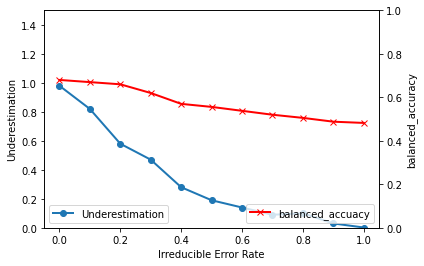}
\caption{Decision Tree}\label{DT_IrreducibleError}
\end{subfigure}
\begin{subfigure}[b]{.45\linewidth}
\includegraphics[width=\linewidth]{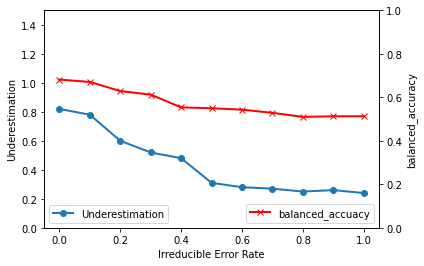}
\caption{k-Nearest Neighbor}\label{KNN_IrreducibleError}
\end{subfigure}

\begin{subfigure}[b]{.45\linewidth}
\includegraphics[width=\linewidth]{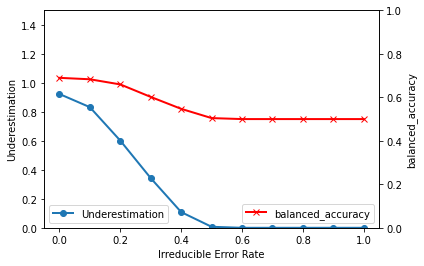}
\caption{Logistic Regression}\label{LR_IrreducibleError}
\end{subfigure}
\begin{subfigure}[b]{.45\linewidth}
\includegraphics[width=\linewidth]{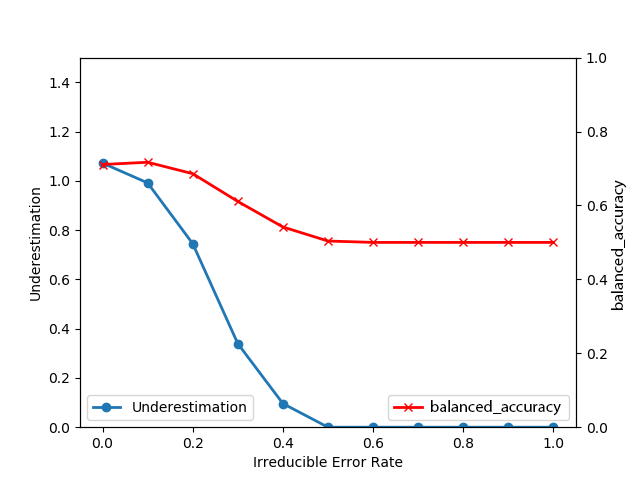}
\caption{Naive Bayes}\label{NB_IrreducibleError}
\end{subfigure}

\begin{subfigure}[b]{.45\linewidth}
\includegraphics[width=\linewidth]{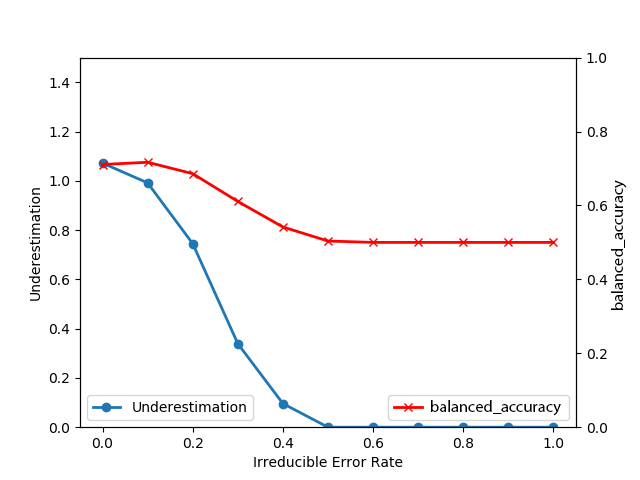}
\caption{Neural Network}\label{NN_IrreducibleError}
\end{subfigure}
\begin{subfigure}[b]{.45\linewidth}
\includegraphics[width=\linewidth]{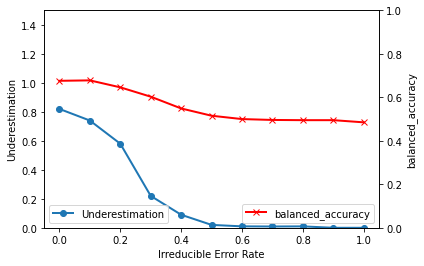}
\caption{Random Forest}\label{RF_BER}
\end{subfigure}
\begin{subfigure}[b]{.5\linewidth}
\includegraphics[width=\linewidth]{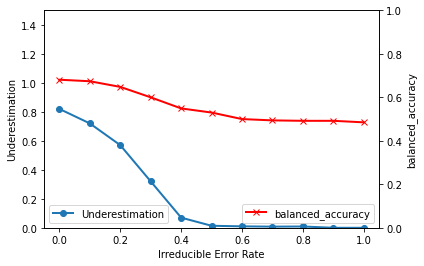}
\caption{Gradient Boosting}\label{GDBT_IrreducibleError}
\end{subfigure}
\caption{Impact of irreducible error on underestimation. }
\label{fig:IrreducibleError}
\end{figure}

\subsection{Impact of Irreducible Error}\label{sec:imp_IrrErr}

To illustrate impact of irreducible error on underestimation, we added Gaussian noise
$X  \sim \mathcal{N}(\mu,\sigma^2)$ to the numeric independent variables. The rationale is that the more noise added, the higher the irreducible error. In this experiment, all seven classifiers are optimized on balanced accuracy with 10-cross validation. The plots in Figure \ref{fig:IrreducibleError} show how underestimation varies as the noise is increased. We see that underestimation is magnified when irreducible errors are high. In addition, we can observe that there is a clear relationship between underestimation and balanced accuracy. This supports our hypothesis presented in section \ref{sec:factors} that high irreducible error results in simpler models, resulting in bias accentuation for the minority class (i.e. bias towards majority).

\subsection{Impact of Regularization}\label{sec:imp_Reg}
\begin{figure}
     \centering
     \begin{subfigure}[b]{0.45\textwidth}
         \centering
         \includegraphics[width=\textwidth]{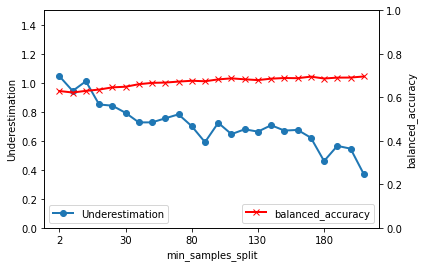}
         \caption{Decision Tree}
         \label{fig:DT_regularization}
     \end{subfigure}
     \hfill
     \begin{subfigure}[b]{0.45\textwidth}
         \centering
         \includegraphics[width=\textwidth]{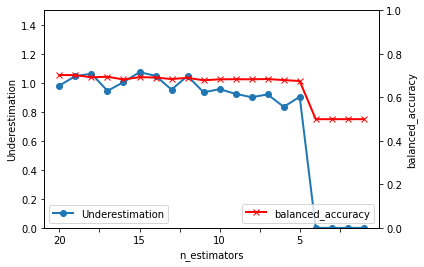}
         \caption{Gradient Boosting}
         \label{fig:GBDT_regularization}
     \end{subfigure}
     \hfill
     \begin{subfigure}[b]{0.45\textwidth}
         \centering
         \includegraphics[width=\textwidth]{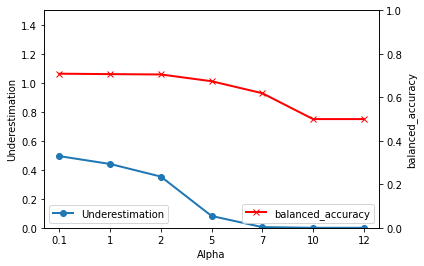}
         \caption{Neural Network}
         \label{fig:NN_regularization}
     \end{subfigure}
     \hfill
     \begin{subfigure}[b]{0.45\textwidth}
         \centering
         \includegraphics[width=\textwidth]{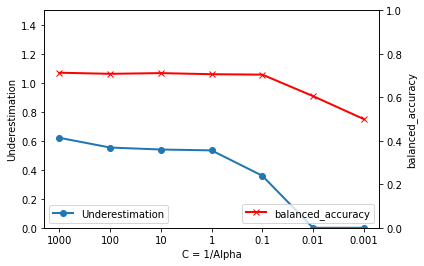}
         \caption{Logistic Regression}
         \label{fig:LR_regularization}
     \end{subfigure}
        \caption{Impact of regularization on underestimation.}
        \label{fig:impactofregularization}
\end{figure}

Next, we evaluated the impact of regularization on underestimation. For this we consider the four classifiers that allow for explicit control of regularization, these are Decision Tree, Gradient Boosting, Neural Network and Logistic Regression. 
For example, the Neural Network implementation provides an $\alpha$ parameter to control over-fitting using $l2$ regularization. Figure \ref{fig:impactofregularization} shows how underestimation varies with these parameters. It is clear that underestimation is exacerbated as the strength of regularization increases (i.e. underfitting). This makes sense since higher regularization means that the model's ability to learn from the training data is restricted, and thus when the observations for minority group are extremely scarce, the model will tend to focus on the majority, resulting in underestimation for minority group. 

\begin{figure}
\centering
\begin{subfigure}[b]{.42\linewidth}
\includegraphics[width=\linewidth]{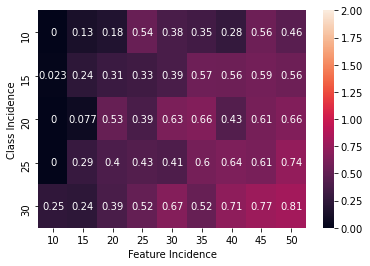}
\caption{k-Nearest Neighbor}\label{knn_classfeatureimb}
\end{subfigure}
\begin{subfigure}[b]{.42\linewidth}
\includegraphics[width=\linewidth]{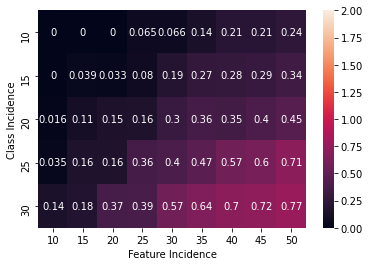}
\caption{Logistic Regression}\label{logreg_classfeatureimb}
\end{subfigure}
\begin{subfigure}[b]{.42\linewidth}
\includegraphics[width=\linewidth]{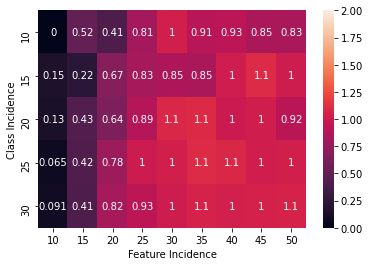}
\caption{Naive Bayes}\label{nb_classfeatureimb}
\end{subfigure}
\begin{subfigure}[b]{.42\linewidth}
\includegraphics[width=\linewidth]{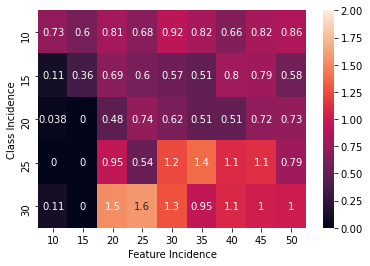}
\caption{Decision Tree}\label{tree_classfeatureimb}
\end{subfigure}
\begin{subfigure}[b]{.42\linewidth}
\includegraphics[width=\linewidth]{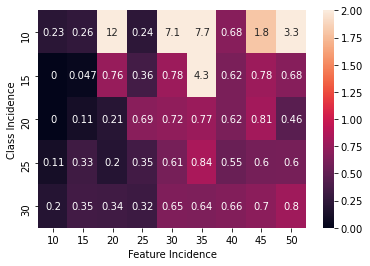}
\caption{Gradient Boosting}
\end{subfigure}
\begin{subfigure}[b]{.42\linewidth}
\includegraphics[width=\linewidth]{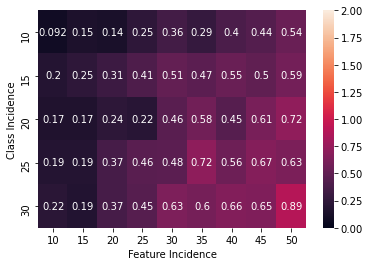}
\caption{Random Forest}
\end{subfigure}
\begin{subfigure}[b]{.42\linewidth}
\includegraphics[width=\linewidth]{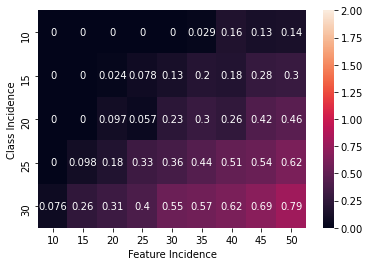}
\caption{Neural Network}
\end{subfigure}
\caption{Impact of class and feature imbalance on underestimation.}\label{fig:ClassFeatureImb}
\end{figure}

\subsection{Impact of Class and Feature Imbalance}\label{sec:imp_Imb}
Lastly, we illustrate how sensitive underestimation is to distribution variations in the class label and the sensitive feature. As can be seen in the heatmaps in Figure \ref{fig:ClassFeatureImb}, the Class Imbalance $P(Y = 1)$ varies between 10\% and 30\% and the Feature Incidence $P(Y = 1|S = 0)$ between 10\% and 50\%. 
Each model is optimized on balanced accuracy with 10-fold cross-validation. We see that the color of the heatmap gets darker as it gets to the most extreme case of class and feature imbalance  (top left corner). This indicates that underestimation is accentuated when observations for minority class is scarce and categories are extremely underrepresented. 

\section{Remediation}\label{sec:fix}

In this section, we discuss possible strategies to remediate underestimation. We have shown in Section \ref{sec:syn_data} that underestimation occurs when the classifier underfits the data due to a combination of limitations in training data and model capacity issues. Hence, the most obvious approach to fix this problem is to get more data. Unfortunately, quantifying exactly how much data is needed is not a trivial task in practice. 

Many of the strategies for addressing Disparate Impact are also applicable for this more specific task of addressing underestimation, these include:
\begin{itemize}
    \item \textbf{Pre-processing:} increasing the sample size for minority group, 
      \item \textbf{In-processing:} adding a constraint to a specific algorithm's loss function to account for underestimation, using cost-sensitive learning, or explicitly considering underestimation in hyper-parameter tuning, 
    \item \textbf{Post-processing:} or selecting different optimal threshold value for the minority group.
\end{itemize}

We evaluated multiple strategies to reduce the impact of underestimation. First, we explicitly consider underestimation in hyper-parameter tuning. However, preliminary experiments on this strategy did not work out, showing only 2\% improvement in underestimation. 
We have had more success with pre-processing strategies and we report on these results here. We have evaluated two strategies for generating counterfactuals (see section \ref{sec:AddingCFs}) and a variation on the SMOTE algorithm for generating synthetic data samples. 

SMOTE \cite{10.5555/1622407.1622416} is perhaps the most popular method for addressing class imbalance in ML. SMOTE is a data augmentation strategy whereby synthetic samples of the minority class are generated by interpolating between real samples. Our modification on this (SMOTE$_F$) is to only select synthetic samples corresponding to the minority group. These are referred to as $S_0Y_1$ in Algorithm 1, i.e. the Discriminated feature and the Positive class. 
In the notation of Algorithm 1, the Favoured Positive ($S_1Y_1$) samples produced by SMOTE are discarded. 

\subsection{Adding Counterfactuals}\label{sec:AddingCFs}

\begin{figure}
     \centering
     \begin{subfigure}[b]{0.45\textwidth}
         \centering
         \includegraphics[width=\textwidth]{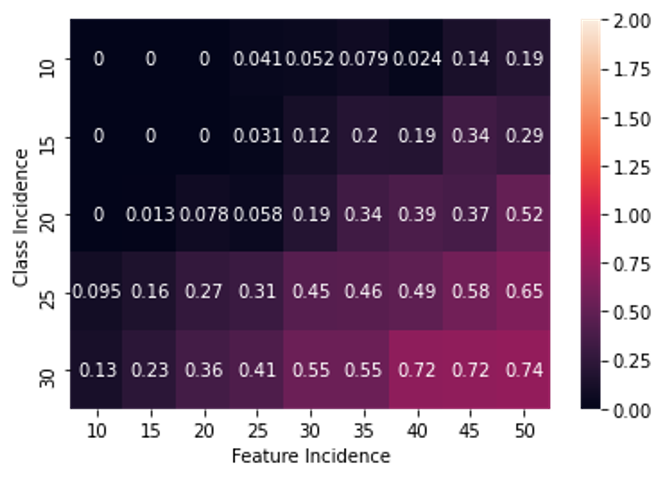}
         \caption{No repair}
     \end{subfigure}
     \hfill
     \begin{subfigure}[b]{0.45\textwidth}
         \centering
         \includegraphics[width=\textwidth]{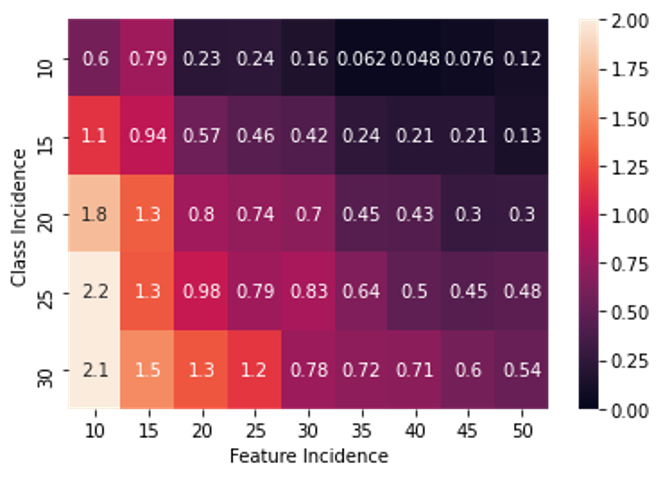}
         \caption{Smote$_F$}
     \end{subfigure}
     \hfill
     \begin{subfigure}[b]{0.45\textwidth}
         \centering
         \includegraphics[width=\textwidth]{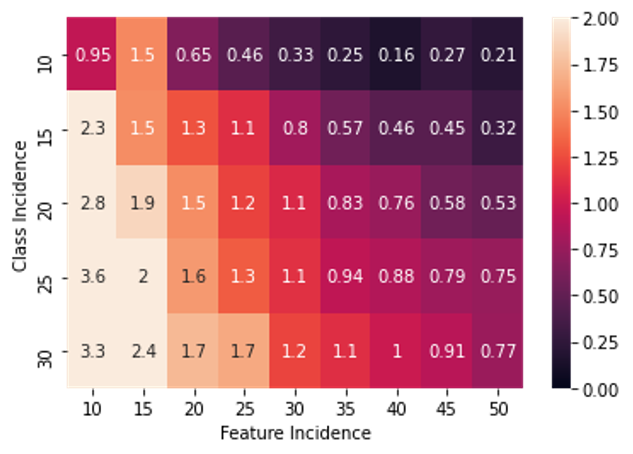}
         \caption{Counterfactual$_F$}
     \end{subfigure}
     \hfill
     \begin{subfigure}[b]{0.45\textwidth}
         \centering
         \includegraphics[width=\textwidth]{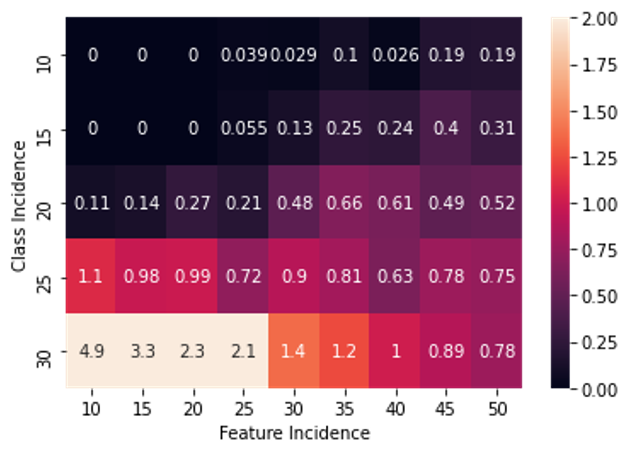}
         \caption{Counterfactual$_L$}
     \end{subfigure}
        \caption{The impact of adding synthetic samples using Smote$_F$ and the two counterfactual strategies. In this baseline analysis we always double the size of the $DP$ set (see Algorithm 1).}
        \label{fig:counterfactuals_synthetic}
\end{figure}

Counterfactual reasoning is an idea from Philosophy that first received attention in Artificial Intelligence research in the 1980s \cite{ginsberg1986counterfactuals}. More recently counterfactuals have been used for explanation and data augmentation in ML \cite{keane2020good,NIPS2017_a486cd07}. Whereas our SMOTE$_F$ strategy generates synthetic examples by interpolating between real ($Y = 1|S = 0$) examples
we have two options for creating counterfactuals: 
\begin{itemize}
    \item \textbf{Counterfactual$_F$}: creating a desirable outcome for the minority group ($Y = 1|S = 0$) from a undesirable outcome ($Y = 0|S = 0$), or
    \item \textbf{Counterfactual$_L$}: a desirable outcome for the minority group ($Y = 1|S = 0$) from one of the majority group ($Y = 1|S = 1$). 
\end{itemize}
The pseudocode for these strategies is shown in Algorithm \ref{adding_counterfactuals_label}. A key consideration with the three data augmentation strategies is the number of new samples $N$ required to mitigate underestimation ($\mathrm{US}_{S} = 1$). In our first experiments we simply double the number of samples representing the desirable outcome for the minority group ($Y = 1|S = 0$). The results on varying  levels  of under-representation of class labels and sensitive features are shown in Figure \ref{fig:counterfactuals_synthetic}. The classifier is the Neural Network used in earlier experiments. 
Due to the stochastic nature of the counterfactuals generation process, we repeat each experiment twenty times and the median of underestimation is obtained.

\begin{algorithm}
    \SetKwInOut{Input}{Input}
    \SetKwInOut{Output}{Output}
    \Input{A dataset $D (X,Y,S)$, classifier $h(x) \rightarrow[0,1]$ }
    \Output{Classifier trained on repaired dataset $h(\Bar{D}) \rightarrow [0,1]$}
    \begin{enumerate}
        \item Divide $D$ into four groups such that:
        \subitem $S_0Y_1 \leftarrow \{x \in D  |  x.S = 0 \wedge x.Y = 1$\}
        \subitem $S_0Y_0 \leftarrow \{x \in D  |  x.S = 0 \wedge x.Y = 0$\}
        \subitem $S_1Y_1 \leftarrow \{x \in D  |  x.S = 1 \wedge x.Y = 1$\}
        \subitem $S_1Y_0 \leftarrow \{x \in D  |  x.S = 1 \wedge x.Y = 0$\} 
        \item  $N \leftarrow |S_0Y_1|$
        \item if Counterfactual$_L$:
        \subitem Randomly sample $N$ observations from $S_0Y_0$ with replacement
        \subitem Change the class labels of the sampled data $Y$ to 1
        \newline
        elif Counterfactual$_F$:
        \subitem Randomly sample $N$ observations from $S_1Y_1$ with replacement
        \subitem Change the sensitive attribute of the sampled data $S$ to 0
        \item Concatenate $D$ with the sampled data
        \item Train a classifier $h(x)$ on modified $D$
        \item Return Classifier trained on repaired dataset $h(\Bar{D}) \rightarrow [0,1]$
    \end{enumerate}
    \caption{Adding counterfactuals}
    \label{adding_counterfactuals_label}
\end{algorithm}

 We can see that both Counterfactual$_F$ Counterfactual$_L$ performs best in almost all cases of under-representation. In fact, both of these strategies overshoot the underestimation target ($\mathrm{US}_{S} = 1$) in extreme cases. In the next subsection we use a cross-validation strategy on the training data to select appropriate values for $N$.

\subsection{Tuning}\label{sec:tuning}
It is clear from Figure \ref{fig:counterfactuals_synthetic} that the impact of data augmentation varies considerably depending on the dataset make-up. For instance, the Counterfactual$_F$ strategy produces $US_S$ outcomes that vary between 0.21 and 3.3. However, we should be able to tune the process by estimating the best value for $N$ using cross-validation.

\begin{figure}
    \centering
    \includegraphics[width=0.65\textwidth]{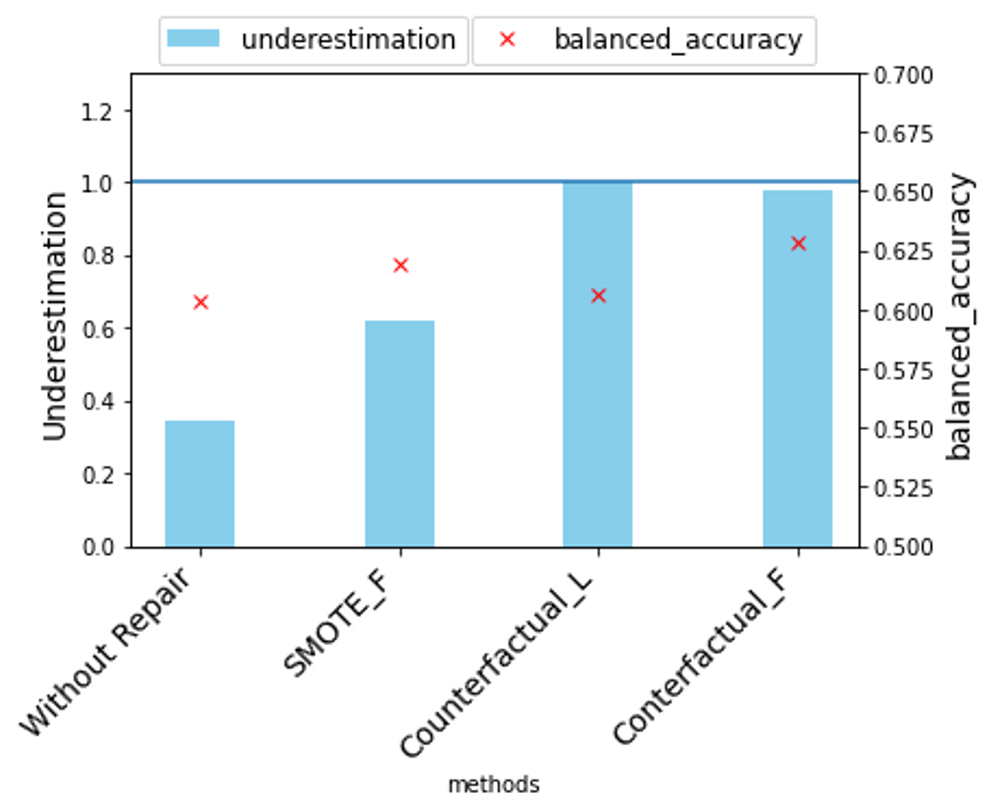}
    \caption{Tuned remediation: Remediation is repeated on the dataset from Figure \ref{fig:counterfactuals_synthetic} with class imbalance of 20\% and feature imbalance of 45\% but with $N$ set using cross-validation on the training data.}
    \label{fig:tunedFix}
\end{figure}

Given that Counterfactual$_L$ draws from $S_0Y_0$ and Counterfactual$_F$ draws from $S_1Y_1$ we can use up to 100\% of these sets in the augmentation process. The validation process (i.e. parameter tuning) considers options from 5\% to 100\% in steps of 5\%. In this experiment the SMOTE$_F$ strategy is also tuned by considering options from 5\% to 100\% of $N \leftarrow |S_1Y_1| - |S_0Y_1|$.


We used a cross-validation process to select the best value for $N$ for each strategy, then the whole training dataset is augmented and the remaining 30\% that has been held back for testing is used to assess performance. The results are shown in Figure \ref{fig:tunedFix}. The tuned SMOTE$_F$ does little to improve underestimation but the two counterfactual strategies do a good job, both bringing $US_S$ close to 1.0. It is worth noting that, in addition to fixing underestimation, Counterfactual$_F$ also improves balanced accuracy from 60.3\% to 62.8\%. 

\section{Underestimation on Real Datasets} \label{sec:real_data}

In this section, we experimentally validate our proposed remediation strategies described in Section \ref{sec:fix} on the Census Income dataset \cite{kohavi1996scaling} and a reduced version of the ProPublica Recidivism dataset \cite{dressel2018accuracy}. These datasets have been extensively studied in fairness research because there is clear evidence of negative legacy. Summary statistics for these datasets are provided in Table \ref{tab:datasets}.
For the Census Income dataset the prediction task is to determine whether a person earns more or less than \$50,000 per year based on their demographic information. The reduced and anonymized version of the Recidivism dataset includes 7 features and the target variable represents whether a person got rearrested within two years after the first arrest. The goal of our experiment is to learn an underestimation-free classifier while maintaining high balanced accuracy score when \textit{Sex} and \textit{Caucasian} used as sensitive feature $S$ for Income and Recidivism dataset, respectively. The classifier is the Neural Network as used in section \ref{sec:tuning} and the same tuning strategy is employed to determine the level of data augmentation.

\begin{table}
\begin{center}
\caption{Summary details of the Adult and Recidivism datasets.}\label{tab:datasets}
\begin{tabular}{ l | r | c| c}
Dataset & Samples & Features & \% Minority \\
\hline
Census Income & 48,842 & 14 & 25\% \\
Recidivism & 7,214 & 7& 45\% \\
\end{tabular}
\end{center}
\end{table}


\begin{figure}
     \centering
     \begin{subfigure}[b]{0.49\textwidth}
         \centering
         \includegraphics[width=\textwidth]{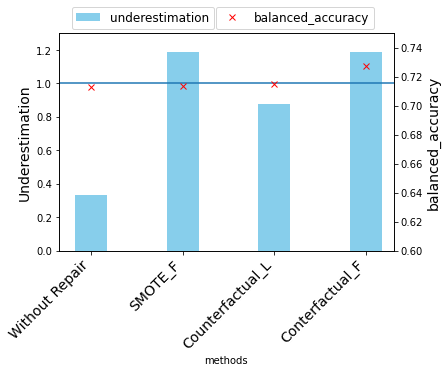}
         \caption{Census Income}
         \label{fig:adult}
     \end{subfigure}
     \hfill
     \begin{subfigure}[b]{0.49\textwidth}
         \centering
         \includegraphics[width=\textwidth]{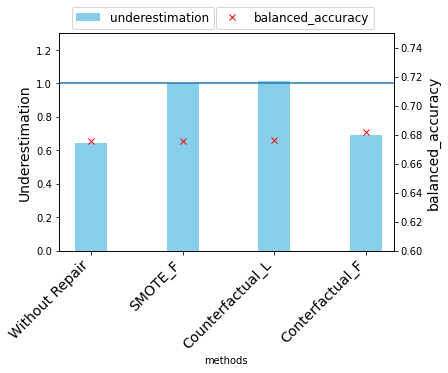}
         \caption{Recidivism}
         \label{fig:recidivism}
     \end{subfigure}

        \caption{An evaluation of the remediation strategies on two real-world datasets.}
        \label{fig:remediationstrategy}
\end{figure}

The results shown in Figure \ref{fig:remediationstrategy} demonstrate that all of the proposed methods reduce the impact of underestimation. Looking closer at the Census Income dataset, we see that Counterfactual$_F$ overshoots underestimation by \~ 10\%. We suspect that this impact can be removed (or at least reduced) by allowing larger search space for $N$ and hence, we can conclude that the proposed strategies are very sensitive to hyper-parameter $N$. Moreover, the results in Figures \ref{fig:tunedFix} and \ref{fig:remediationstrategy} show that there is no one remediation strategy that performs best in all datasets, suggesting the importance of tuning. It is worth noting that Counterfactual$_F$ improves balanced accuracy for all three datasets. This suggests that it is a natural policy for producing counterfactuals. 

These methods could also easily be extended to account for multi-label sensitive group. We believe that the effectiveness of these strategies is dependent on how different the distribution of the test set to the training set. More research need to be done to further refine these methods to account for distribution shift.

\section{Conclusions and Future Work}
This paper started by emphasising the difference between negative legacy and underestimation as sources of bias in ML. The first is a problem with the data, the second is a problem with the algorithm. We have shown that underestimation can be addressed by adding counterfactuals, suggesting that perhaps it is really all about the data. It's not. Negative legacy refers to scenarios where there are undesirable patterns in historic data. Underestimation refers to scenarios where the algorithm is not picking up the patterns that \emph{are} in the data. It is worth saying then that fixing underestimation in the sense that the algorithm reflects the data may still leave a fairness problem due to negative legacy. 

While our evaluation shows that the use of counterfactuals can ameliorate underestimation there is still some room for improvement. Our next step will be to use ideas from the work of Keane and Smyth \cite{keane2020good} to select better counterfactuals. In future we plan to move on to develop in-processing strategies whereby underestimation is explicitly considered in the ML algorithm optimization process.

\section*{Acknowledgements}
This work was funded by Science Foundation Ireland through the SFI Centre for Research Training in Machine Learning (Grant No. 18/CRT/6183) with support from Microsoft Ireland.

\bibliographystyle{splncs04}
\bibliography{bias.bib}

\end{document}